# Artificial intelligence, rationalization, and the limits of control in the public sector: the case of tax policy optimization

Jakob Mökander[1,2], Ralph Schroeder[1,3]

[1] Oxford Internet Institute, University of Oxford, 1 St Giles, Oxford OX1 3JS, UK
[2] Center for Information Technology Policy, Princeton University, Princeton, NJ 08544, US
[3] Center for Advanced Study in the Social Sciences, Stanford University, Stanford, CA 94305, US

Email for correspondence < jakob.mokander@outlook.com >

**Abstract**

In this paper, we first frame the use of artificial intelligence (AI) systems in the public sector as a continuation and intensification of long-standing rationalization and bureaucratization processes. Drawing on Weber, we understand the core of these processes to be the replacement of traditions with instrumental rationality, i.e., the most calculable and efficient way of achieving any given policy objective. Second, we demonstrate how much of the criticisms, both among the public and in scholarship, directed towards AI systems spring from well-known tensions at the heart of Weberian rationalization. To illustrate this point, we introduce a thought experiment whereby AI systems are used to optimize tax policy to advance a specific normative end: reducing economic inequality. Our analysis shows that building a machine-like tax system that promotes social and economic equality is possible. However, our analysis also highlights that AI-driven policy optimization (i) comes at the exclusion of other competing political values, (ii) overrides citizens' sense of their (non-instrumental) obligations to each other, and (iii) undermines the notion of humans as self-determining beings. Third, we observe that contemporary scholarship and advocacy directed towards ensuring that AI systems are legal, ethical, and safe build on and reinforce central assumptions that underpin the process of rationalization, including the modern idea that science can sweep away oppressive systems and replace them with a rule of reason that would rescue humans from moral injustices. That is overly optimistic: science can only provide the means – they cannot dictate the ends. Nonetheless, the use of AI in the public sector can also benefit the institutions and processes of liberal democracies. Most importantly, AI-driven policy optimization demands that normative ends are made explicit and formalized, thereby subjecting them to public scrutiny, deliberation, and debate.

## 1. Introduction

The use of artificial intelligence (AI) systems increasingly permeates modern societies, including the public sector (Wirtz et al., 2019). This means that many decisions that were previously made by human experts are now made by AI systems (Zarsky, 2016). The drivers behind this development are clear: enabled by recent advances in machine learning (ML) research and fuelled by the growing availability of large, fine-grained digital data sources (Wiggins & Jones 2023), AI systems can improve the efficiency and consistency of existing decision-making processes and enable new solutions to complex optimization problems (Taddeo & Floridi 2018). AI systems enable good governance in two ways: directly, by automating and personalizing public service delivery, and indirectly, by informing policy design through more accurate forecasting and simulations of complex systems (Margetts & Dorobantu, 2019). These advantages are not merely hypothetical. Empirical studies have documented governments' use of AI systems across a range of applications and geographic settings; from streamlining immigration procedures through biometric and biographic matching in New Zealand (Nalbandian, 2022) to facilitating preventive healthcare in China (Sun & Medaglia, 2019).

However, the use of AI systems in the public sector is coupled with ethical and legal risks. For example, it may produce discriminatory outcomes, violate individual privacy, and enable human-wrongdoing (Tsamados et al., 2021). A study of COMPAS – an AI-powered decision tool used by US courts to predict criminal recidivism – found that it systematically discriminated against African-American defendants by overestimating their likelihood of reoffending (Angwin et al., 2016). Another controversy took place in the Netherlands, this time surrounding a data-driven welfare fraud detection system referred to as 'SyRI.' In addition to discriminating against minorities, it was found that SyRI's linking of personal data from multiple sources did not comply with the right to privacy under the European Convention on Human Rights (van Bekkum & Borgesius, 2021). In short, there are many examples of instances where AI systems have caused harm, or at least failed to work as advertised (Kapoor & Narayanan, 2022).

Against this backdrop, it is understandable that researchers and policymakers have called for increased fairness, accountability, and transparency with respect to the use of AI systems. There is, for example, a vast academic literature on how to design and deploy AI systems that are legal, ethical, and safe.[1] Standard-setting bodies have provided similar guidance[2] and policymakers have proposed hard regulations to manage the risks AI systems pose.[3] These initiatives tend to emphasize the novelty of AI systems and the ethical challenges they pose. The upside of this framing is that it gives a sense of urgency to the cause of addressing the issues under consideration here. However, there are also downsides to overemphasizing the novelty of specific technologies and social phenomena.

In this paper, we argue that the contemporary discourse concerning the social challenges associated with the use of AI in the public sector is best understood against the backdrop of a longer trajectory whereby calculability is increasingly imposed on social processes. Specifically, we demonstrate that the use of AI systems in the public sector can be viewed as a continuation and intensification of ongoing rationalization and bureaucratization processes. Drawing on the work of Max Weber, we take the core of these processes to be the replacement of personalistic rule, tradition, and emotion as motivations for behaviour by *instrumental rationality*, i.e., the most calculable, efficient, predictable, and impersonal way of achieving any given policy objective. The reader will notice that the characteristics of rationalization correspond to the traditional goals of AI research, e.g., perception, learning, knowledge representation, reasoning, and planning (Russell & Norvig, 2016).

---

[1] See, e.g., Russell (2019), Floridi et al. (2018) or Laufer (2022).
[2] Examples include *Ethically Aligned Design* (IEEE, 2019) and NIST's (2023) *AI Risk Management Framework*.
[3] In May 2021, the European Commission published a draft version of the *AI Act*. A similar bill, the *Algorithmic Accountability Act of 2022*, is currently being discussed in the U.S. Congress (Mökander & Floridi, 2022).

The advantage of foregrounding this perspective is that addressing a specific problem requires both an adequate conceptualization of the phenomenon in question and the mechanisms driving it and a realistic and pragmatic understanding of scientific knowledge (Hacking, 1983). Since much of the public outcry and scholarly criticism directed towards "AI" mirrors well-known critiques of rationalization formulated by Weber, viewing the use of AI systems in the public sector as the latest stage in this long-standing yet accelerating trend of technoscientific rationalization allows us to draw on a rich and established vocabulary to better articulate and diagnose specific problem formulations.

It is true that AI systems pose a wide range of ethical and legal risks. It is also true that many AI systems fail to live up to the claims made by the organizations that deploy them. In these cases, more robust software development practices (Kearns & Roth, 2020), broader impact requirements (Prunkl et al., 2021), guardrails for the use of AI systems (Gasser & Mayer-Schoenberger 2024), and independent AI audits (Mökander, 2023) can help mitigate risks and prevent harms. However, with respect to the cold, impersonal treatment of decision subjects (O'Neil, 2016), the quantification of social relationships (Mau, 2020), the centralization of decision power and the standardization of decision criteria (Kleinberg et al., 2021), the automation of cognitive tasks (Brynjolfsson & McAfee, 2014), or the shaping of social preferences through nudges (Thaler & Sunstein, 2008), the problem is not that AI systems "don't work" but that they work "too well" as tools for instrumental rationalization.

The argument of this paper can thus be summarised in two claims. First, the social and ethical challenges associated with AI systems intensifying ongoing rationalization and bureaucratization processes are distinct from those resulting from AI systems failing to work as advertised. Second, these two distinct sets of social and ethical challenges call for different remedies, which sometimes stand in tension with each other.

This line of reasoning is best illustrated by an example. In this paper, we examine how AI systems could be employed to achieve specific normative ends through policy optimization. Building on recent work in economics (Kasy, 2018), political philosophy (Sparks & Jayaram, 2022), and computer science (Zheng et al., 2022), we propose a thought experiment whereby the state uses AI systems to not only *automate* the tax system but also to *optimize* tax policy in order to reduce economic inequality. While our analysis shows that such uses of AI systems are within the realms of possibility, it also surfaces the social and ethical tensions associated with rationalization and bureaucratization processes. Specifically, our thought experiment highlights how AI-driven policy optimization may not only be in tension with how consensus over competing political values is reached in democratic societies but may also override citizens' senses of their (non-instrumental) obligations to each other as well as their self-understandings as self-determining beings.

The remainder of this paper is structured as follows. In Section 2, we frame the use of AI systems in public administration as a continuation and intensification of rationalization and bureaucratization processes that have been going on for several centuries. In Section 3, we review relevant literature on tax policy and how it can be used to pursue certain ideas about equality using knowledge that tries to reshape peoples' preferences and behaviour. In Section 4, we detail our thought experiment about how to reshape equality through AI-driven tax policy optimization. In Section 5, we discuss the social and ethical implications of that experiment. In Section 6, we conclude the paper by generalizing our argument and highlighting how AI driven policy optimization – despite the tensions it causes –can also benefit the institutions and processes of liberal democracies by demanding that normative ends are made explicit and formalized, thereby subjecting them to public scrutiny and debate.

## 2. Artificial intelligence as Weberian rationalization

In this section, we define the two central concepts of our analysis: AI and rationalization. Both are disputed concepts which have eluded widely agreed-upon definitions. For example, Legg and Hutter (2007) list no fewer than 70 competing and partly conflicting definitions of AI. Similarly, the term rationalization has different meanings in different contexts.

Following the OECD (2021), we define an AI system as a machine that can, for a given set of objectives, make predictions, recommendations, or decisions influencing its environment. It does so by processing input data to (i) perceive its environment, (ii) abstract perceptions into models of the external environment, and (iii) use model inference to structure information or perform actions in an automated manner. An AI system "learns" insofar as it updates its model and internal decision-making logic as it is fed new input data. The model itself can be based on ML algorithms such as deep neural networks (LeCun et al., 2015), formal logic as in expert systems (Giarratano & Riley, 2004), or a combination thereof as in hybrid systems (Marcus, 2020). In some cases, a model based on deterministic rules offers a single recommendation. In other cases, models based on probabilistic reasoning can offer a variety of recommendations. In each case, however, pre-defined objectives stipulated by humans guide both the model inference phase and the subsequent execution.

For our purposes, this definition has three advantages. First, it bypasses any distracting discussions about machine consciousness by focusing solely on the observable characteristics and operations of AI systems. As noted by Esposito (2022), if AI systems appear intelligent, this is not because they have learned how to think but because we have learned how to communicate with them in ways that advance our purposes. Second, the definition is broad enough to encompass symbolic (logic-based) and sub-symbolic (ML-based) AI systems, both of which play important roles in optimizing social policies and automating their execution. Third, it highlights how AI systems operate with varying levels of complexity and autonomy in larger human-centric decision-making processes (Mökander et al., 2023). To explain why that is the case, we must first define our second central concept.

With *rationalization*, Weber understood a long-term process whereby beliefs based on tradition are replaced by rules based on logic and (instrumental) efficiency (Brubaker, 1984). As Weber observed, the process of rationalization takes different forms in the economic, political, and cultural spheres of life (Gellner, 1992), which makes it difficult to summarize neatly. But we focus here on politics and how the state imposes calculability on social processes. As we shall see, this is where Weber's ideas fit closely with how AI systems can both automate and optimize policymaking processes.

In relation to the state, Weber thought that rationalization was synonymous with a machine-like apparatus of rules, i.e., bureaucratization. The modern state for Weber thus rests on the legitimacy of legal authority or domination via a system of predictable and impersonal rules (Mommsen, 1984; Breuer, 1998). Legal-rational authority is embodied in bureaucracy and closely related to instrumental rationality since it contrasts with traditional or charismatic authority. The advantage of this form of rule is equal treatment and greater efficiency. The disadvantage, in Weber's view, is his famous "iron cage", i.e., the inescapability of structures that constrain freedom because they leave less room for individual autonomy (Baehr, 2001). Weber's idea of history was rather pessimistic on this point. He thought that the disenchantment of the modern world went hand in hand with ever-greater constraints on freedom. But it is not necessary to follow his pessimism: a well-working state bureaucracy can also allow more space to lead one's private life in an unfettered way.

The point to stress from Weber's theory of rationalization is that computation can be viewed as a ubiquitous phenomenon not limited to AI systems or other technical artefacts. As he puts it, the "peculiar modern Western form of capitalism [is]…strongly influenced by its technological conditions. Its rationality is today essentially dependent on the calculability of the most important technical factors. But this means fundamentally that it is dependent on the peculiarities of modern science, especially the natural sciences based on mathematics and exact and rational experiment" (Weber, 1967: 24). Weber's "disenchantment" is thus based on the idea that "one can, in principle, master all things by calculation" (1948: 139) – and computation by ML-based AI systems is perhaps the prime current instantiation of that mastery.

We are not the first to observe the link between the use of AI systems in the public sector and long-standing rationalization processes. In *Digital Weberianism*, Muellerleile and Robertson (2018) argue that, far from constituting a radical rupture, the use of AI systems and data in today's digitizing society show strong traces of the logic of Weber's bureaucracy. Similarly, Vogl et al. (2020) document British local governments' use of AI systems to describe the emergence of what they call an *algorithmic bureaucracy*. These works have provided a theoretical framework on which our analysis builds. Yet our contribution moves beyond previous work in two ways. First, we explore the implications of rationalization in a specific domain: AI-driven tax policy optimization. Second, we highlight the normative tensions associated with AI rationalization and spell out their implications for the ongoing discourse on fairness, accountability, and transparency in AI research and system design.

Before proceeding, it should be noted that AI systems facilitate the operations of the bureaucratic state in two conceptually distinct ways. First, they can be used to automate routine tasks, such as diagnosing patients, grading essays, or calculating tax returns. Historically, such tasks have been performed by "street-level bureaucrats" (Lipsky, 1980) who, despite following rules that theoretically applied equally to all citizens, inevitably operated with some degree of discretion. In an article titled *Rule by Automation*, Sparks and Jayaram (2022) argue that AI systems can promote freedom and equality by automating the tasks of street-level bureaucrats. By eliminating human discretionary power, rule by automation reduces our dependence on the intentional and arbitrary will of other actors and reduces social hierarchy. Hence, Sparks and Jayaram conclude, rule by automation should be supported for the same reasons as the rule of law.

Second, AI systems can be used to gather and process information to inform or optimize policies designed to reach specific normative ends. Of course, controlled policy experiments are difficult to conduct (Barnow, 2010). However, bureaucracies already claim to apply "evidence-based policymaking", i.e., the idea that policies should be informed by objective knowledge about the state of the world and the causal effects of specific interventions (Cartwright & Hardie, 2012). This is where AI systems come in. By leveraging the growing availability of fine-grained digital data sources, AI systems can perform large-scale policy experiments in real time and – based on the feedback from such experiments – optimize policy to achieve objectives defined by human policymakers. Figure 1 illustrates the difference between using AI systems to *automate* and *optimize* policymaking.

*Figure 1. The conceptual difference between automating tasks and optimizing processes.*

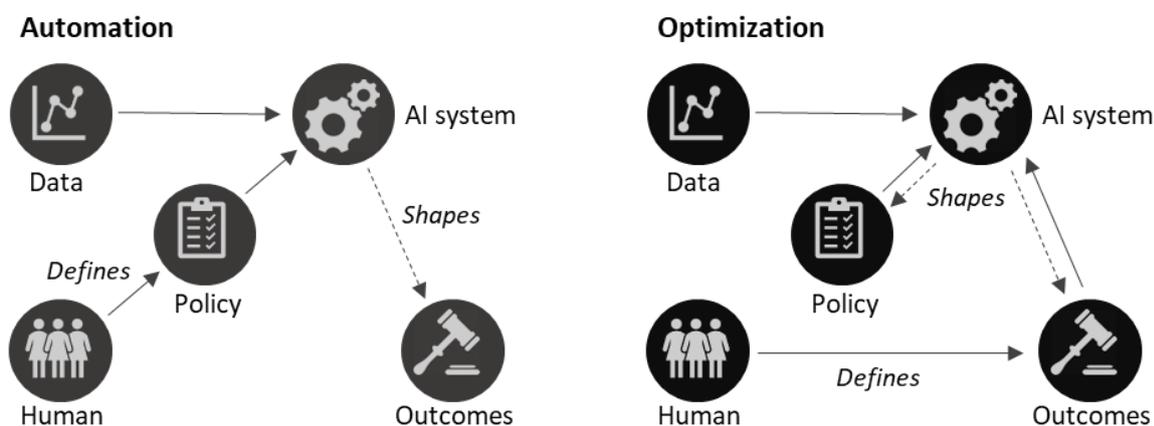

The distinction between AI-driven automation and optimization must not be overemphasized; the two functions both build on and intensify the logic of rationalization inherent in bureaucratic states. Moreover, optimizing certain processes may well involve automating repetitive tasks to reduce errors and improve productivity. However, the two functions give rise to different normative tensions. The tensions associated with AI systems used to automate tasks have sparked much debate, especially when those systems fail. In contrast, the tensions created by using AI systems for policy optimization – even when they work as intended – have received little attention. Our thought experiment in Section 4 will surface those tensions. But first, something should be said about why we focus on AI-driven tax policy as an example.

## 3. Bureaucratization, tax policy, and equality

In this section, we review relevant literature about tax policy. In doing so, we show how the institutional histories of nation-states and the associated civic behaviour of national populations shape each other. We also highlight how taxation – and experiments with tax policy – can be used to achieve different normative ends.

The rationalization of taxation played a central role in the emergence of modern state bureaucracies. The pre-modern state had neither the capacity nor the information necessary to collect taxes directly without using intermediaries (Scott, 1998). Medieval tax systems involved collective obligations and non-monetary quotas – based on rules of thumb or traditions – on towns and villages, which were administered by local nobility. The story of how modern taxation regimes emerged with warfare between modern European states is well-known (Tilly, 1990). Subsequent developments meant that rational taxation rests on explicit and calculable rules that are systematically enforced on whole populations. The point is that sources of revenue have a major impact on patterns of state formation (Moore, 2004), and that states require intermediaries – including information gathering infrastructures and bureaucratic organizations – to collect taxes from their populations).

Following a similar line of reasoning, Weber argued that rational taxation is a way of financing the state that encourages and is encouraged by the expansion of bureaucracies. Because collecting taxes effectively is complex and requires cooperation across state agencies, it required modern states to invest in new capabilities and spearheaded the broader development of bureaucratic institutions (Dandeker, 1990; Besley & Persson, 2009). Today, most developed countries have highly rationalized tax systems supported by large state bureaucracies. However, the relationship between a state's institutional history and the civic behaviour of its population is dynamic (Steinmo, 1993). Citizens who are required to pay taxes are more likely to feel ownership of government activities and to make demands for representation. Similarly, governments in need of tax revenue have stronger incentives to make reciprocal concessions to taxpayers and encourage tax compliance. Consequently, variations in cultural, economic, and political factors in the early modern period influenced not only the features of different countries' tax systems but also the attitudes and behaviours of their populations with respect to taxation.

To illustrate this point, it is useful to consider how taxation works in Sweden and the US, which are often perceived as cases that lie at the extremes of the "varieties of capitalism" (Pontusson, 2005). In *Willingness to Pay*, Steinmo and D'Attoma (2021) demonstrate through empirical research – which includes data from the two countries – that people tend to obey laws and trust institutions more in societies where the rules are clear, coherent, and consistently applied. Paying taxes is disliked in the US partly because it requires a lot of effort to file tax returns and partly because everyone thinks everyone else is getting a better deal from the plethora of tax breaks and exemptions. It can be added that these exemptions are not a matter of tax officials' discretion; rather, the US uses taxation to pursue social policy in various ways, or lawmakers provide exemptions to appeal to their various supporters. In Sweden, the opposite is the case. The Swedish tax system is universal and transparent; everyone can find out how much others are paying and there is little pandering to different groups via deductions. It is also highly efficient; a simple click on one's mobile phone suffices for most people to file tax returns.

This is a good place to consider some reasons why AI systems are likely to be employed to automate taxation and optimize tax policy. To start with, the monetary nature of modern economies and the digitalization of the financial system mean that taxation lends itself to calculation and thus rationalization. As Sweden's case illustrates, data regarding people's income and wealth are readily available in digital form. Further, how tax systems are designed shapes people's attitudes towards paying taxes. As previously stated, people dislike paying taxes when it involves lots of work and they suspect others are getting a better deal. An automated tax system that reduces the scope for arbitrary exemptions would address both concerns. Finally, while few citizens understand complex policy issues like taxation, everyone can understand – and partake in the discussion about – the normative ends that taxation should be directed towards (Steinmo, 1993).

This brings us to the second key point: how tax policy can be used to reshape inequality. Researchers have long studied different tax policies and their effects on economic redistribution (Steinmo, 2010; Hobson, 2003; Prasad, 2018). Mostly, the levelling of income via taxation and public transfers is rather small (Tanzi & Schuhknecht, 2000). As Steinmo (2010:156) observes: "In the United States – as in most countries – the major recipient of public transfers is the middle class." However, Steinmo (2010: 153) also notes that "the US is unique in the extent to which it attempts to regulate, reward, subsidize and manipulate the behaviour of different actors through its tax system rather than through public spending." This means that taxation is one of the levers states can use to reduce economic inequality – provided the tax policy is directed towards that end.

Fairness in relation to taxation is, of course, part of a larger debate about equality. And the public's ideas about equality are far from consolidated. Citizens perceive that taxes are their contribution to society or to each other. While perceptions matter, Prasad (2018: 214) highlights a critical tension:

"Political scientists have noticed that Americans are ideologically conservative but pragmatically liberal – they love the idea of small government in abstract terms, but in practice they like almost everything that government does." Or, as Steinmo (1993:17) puts it: "Citizens want higher public spending *and* lower taxes." In short, popular support for more progressive taxation is mixed at best.

Nevertheless, the political philosophy in relation to the distribution of resources is clear: Marxists view reduced inequality as an intrinsic good (Wolf, 2012); egalitarian utilitarians want to maximize well-being across the population (Singer, 2002); liberals want to maximize the capabilities of citizens (Sen, 2009); and conservatives seek a normative foundation in national identities that transcends class divisions (Scruton, 2014). All these political projects require systematic taxation and the continuous redistribution of resources to reduce the inequalities to which markets give rise, though in very different ways.

Among policymakers, the question of taxation and equality has once again risen to the top of the agenda, as evidenced by the widespread attention given to the work of Thomas Piketty (2014). Piketty has generated much debate about how taxation – and the public spending arising from it – could be engineered to overcome rising inequalities. However, how to design tax systems that reduce inequality is also an empirical question and, as the next section shows, one that is best answered through a data-driven and experimental approach.

To summarize, we focus on the example of tax policy for three reasons. First, taxation lies close to the centre of the bureaucratic state's capabilities. Second, taxation can be used as a governance mechanism to manipulate the behaviour of populations and thereby achieve specific normative ends. Third, taxation is a domain that, by virtue of being readily quantifiable, lends itself well to AI-driven rationalization. In the next section, we will combine these ideas.

## 4. AI-driven tax policy to reduce economic inequality: a thought experiment

In this section, we propose a thought experiment in which AI systems are employed to optimize tax policy and automate taxation to achieve a given normative end: minimizing economic inequality. The aim thereby is to provide an example of how the use of AI systems in the public sector is both a continuation and intensification of Weberian rationalization as well as to highlight the normative tensions to which it gives rise.

Consider the following hypothetical situation: there is a broad political consensus in a specific nation-state regarding the aim of reducing economic inequality, and the policy debate now centres on how that objective can be reached. Admittedly, such a situation is highly unlikely to materialize. Political consensus is rare (Rorty, 2021). Moreover, minimizing inequality is only one conceivable policy goal; others could be envisaged.[4] However, under the *assumption* that there is a broad political consensus to reduce economic inequality, policymakers are left with a problem of mechanism design: how to find a policy under which the affected economic agents yield the desired outcome (Myerson, 1981).

Hitherto, policymakers have approached such a quest by drawing on economic theory regarding taxation (Ramsey, 1927; Mirrlees, 1971; Saez, 2001) and evidence from experimental research on the relationship between policy, civic behaviour, and inequality (Spicer & Becker, 1980; Agranov & Palfrey, 2015). However, while useful, theoretical approaches to policy design are limited since they fail to capture the complexity of the real world (Hayek, 1973). Moreover, although policy experiments have become popular, they are hard to design, and results tend to suffer from limited external validity (Peters et al., 2016). Whether produced by lab or field experiments, it is often unclear how well empirical findings about the effectiveness of specific policies generalize between different social and temporal contexts. Historically, policymakers have thus faced significant uncertainty regarding the actual outcomes of different tax policies.

---

[4] Here, we assume a consensus about egalitarianism as a goal for policy making. In practice, such a goal would and stand in pluralist competition with other goals, such as a Nozickian minimalist state (Nozick, 1974).

This is where AI rationalization comes in. Rather than having humans decide what the policy should be, leaving the actual outcome uncertain, policymakers only need to define the goal function, establish constraints, and employ AI systems to optimize the policy. In Section 2, we defined an AI system as a machine that can, for a given set of objectives, make decisions influencing its environment. It does so in two steps: first, by processing input data to build models of the environment and, subsequently, by using model inference to inform or perform actions in an automated manner. AI systems "learn" insofar as they update their models when fed with new input data. Provided that the goal function is clearly defined, such systems thus offer a fundamentally new approach to policymaking: one that is data-driven and based on learning from continuous feedback.[5]

The idea behind AI-driven policy optimization is straightforward but requires some understanding of reinforcement learning (RL). Simplified, RL is a technique to train AI systems based on neural networks operating in unknown environments to learn the optimal policy (i.e., set of decisions) to maximize a reward function (Sutton & Barto, 2018). As a framework for solving decision problems, RL has been widely employed in areas like natural language processing and medical diagnostics (Mousavi et al., 2018). However, RL is particularly apt for policy optimization for two reasons. First, RL means learning from mistakes. Because RL maintains a balance between *exploration* (trying out what works) and *exploitation* (trying again what has worked in the past), AI systems using RL learn from their own experience over time and do not require all the data needed to solve a problem to be available upfront. Second, RL approaches optimization problems holistically, without dividing them into subtasks. This makes AI systems based on RL well-suited for complex decision problems where the long-term reward is prioritized over short-term benefits.

With this, we can return to our thought experiment. Using RL, an AI system could (i) perform large-scale policy experiments on populations, (ii) observe the effects of different policies in real time, and (iii) continuously revise and refine the tax policy to reduce economic inequality. AI-driven tax policy optimization is thus a good example of Weberian rationalization. Recall that instrumental rationality refers to the most calculable, predictable, and efficient ways of achieving any *given* policy objective. In this case, that objective is to achieve a more equal distribution of resources by finding more effective ways to shape taxpayers' behaviour. Figure 2 illustrates the logic behind this approach.

---

[5] Feedback is a method of controlling a system by reinserting the results of its past performance. As Wiener (1950: 71) put it: If the information which proceeds backward from the performance can change the general method and pattern of performance, we have a process which may well be called learning.

*Figure 2. AI-driven tax policy experimentation and optimization to maximize social equality.*

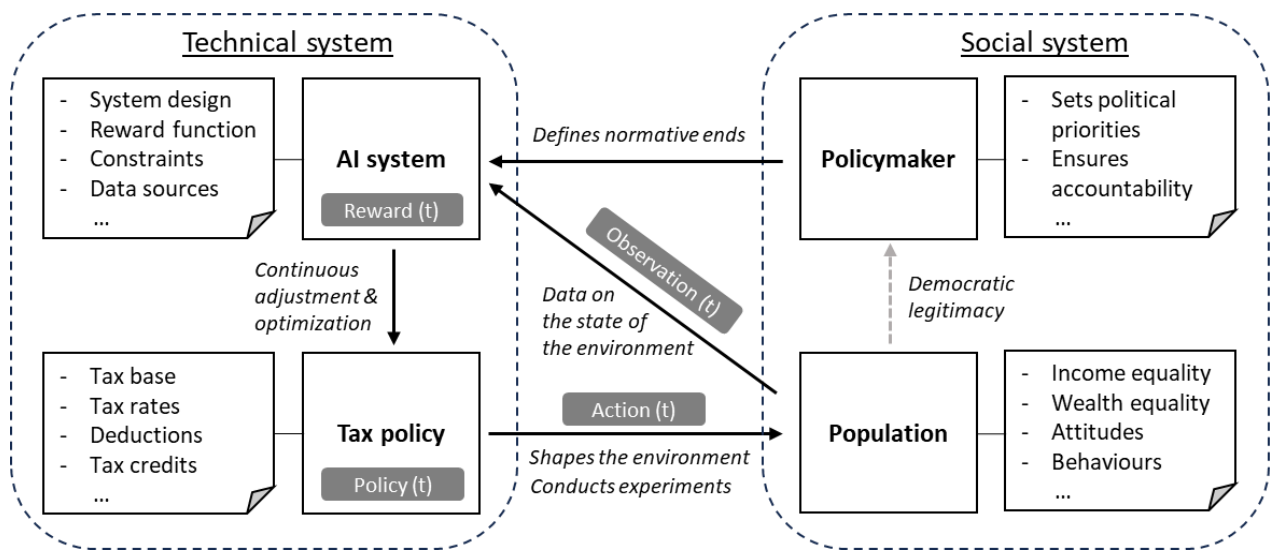

Implementing AI-driven tax policy optimization would require policymakers to make three system design choices upfront. First, policymakers would need to define a goal function. To optimize for economic equality, the goal function could include both (i) *income inequality metrics*, like the Gini index or the Thiel index, and (ii) *wealth inequality metrics*, like the Gini coefficient or the Palma ratio. How to define the goal function is a non-trivial question that lies beyond this paper's scope. However, some general observations can be made. To start with, policymaking is always a multi-variable optimization. Hence, any goal function is likely to consist of a plurality of complementary metrics. Further, equality could be conceived in different ways. Whereas utilitarians and egalitarians focus on ensuring equal outcomes, liberals focus on providing equal capabilities. An AI system employed to maximize equality could equally favour a liberal notion since it is easier to create good records of how much money people have than what they do with it or how it affects their subjective well-being.[6]

Second, policymakers would need to decide on which variables the AI system should be able to manipulate. To maximize its goal function, the AI system would require the ability to manipulate the variables constituting a holistic tax policy, including *tax base* (what should be taxed?), *tax rates* (how should different assets or incomes be taxed?), *tax exemptions* (what deductions should be allowed?), and *tax credits* (what rebates and subsidies should be granted?). However, it should be noted that AI systems based on RL put no value on variables that are not part of the goal function or subject to explicit constraints.[7] Consequently, there are good reasons to limit the number of levers the AI system has control over, e.g., to reduce the likelihood of unintended harms.

Third, policymakers would need to define what data sources the AI system should be able to access. To optimize the tax policy to achieve maximal economic equality, the system would need access to *economic data* (regarding citizens' incomes and wealth), *sociological data* (regarding citizens' attitudes and behaviour) as well as information about how these data change over time. Further, to perform policy experiments, the AI system cannot rely on aggregate data but needs access to fine-grained data on how individuals and groups change their behaviours in response to specific policy treatments.[8] Such experimentation is already a favoured tool in economics (Mascagani, 2017). The difference is that, fed with relevant input data in a seamless and continuous manner, the AI system would conduct large-scale, real-time experiments on the same population for which it optimizes the policy. Therefore, it would overcome the concerns regarding external validity traditionally associated with policy experiments.

---

[6] The goal function would need to include some yardstick or constraint preventing the AI system from aiming at a level of equality that is completely nivellating (Parfit, 1995). Rawls' (1971) maximin principle, which seeks to maximize the welfare of societies' least well-off members, could be one such constraint.
[7] This is what philosophers have referred to as the value alignment problem (Gabriel, 2020). One example is the paperclip maximiser discussed by Bostrom (2014), which highlights the concern that an AI system set to maximize any specific objective would, unless constrained, inadvertently end up causing harm.

Some further clarifications are needed. The key feature of AI-driven policy optimization is *learning*. This means that while the AI system uses data about taxpayers' income, wealth, and behaviour as inputs to calculate the optimal tax policy to minimize economic inequality, this optimum would need to be continually adjusted to account for changing economic and social circumstances. Further, as already indicated, learning is contextual. In Section 3, we demonstrated how each country's institutional history, tax policies, and the civic behaviour of its population shape each other. This means that it is, theoretically, possible to shape the behaviours of populations in the direction of their stated preferences with technocratic solutions. However, it also means that the borders of nation-states would be the main constraint within which learning takes place.

Let us now summarize our thought experiment. We explored the idea of using AI systems to optimize the tax policy of a nation-state to promote economic equality within its population. This would take the form of repeated large-scale experimentation to learn about the behaviour and preferences of taxpayers and the effectiveness and feasibility of different tax policies. This feedback would then be used to continuously adjust the tax policy and shape the behaviour and preferences of the population in line with that policy.[9] This is a critical point, since research has shown that voters often go against their own interests in terms of tax policies (Bartels, 2005). Shaping the preferences of citizens in such a way would mean control through (AI-generated) knowledge, which is fully in line with instrumental rationalization. However, it would mean treating the population as experimental subjects.

Of course, our thought experiment is subject to both technical and political limitations. For example, although it is clear in principle how an AI system could be used to optimize tax policy, many technical challenges, such as data access, could hamper its implementation. AI-driven tax policy optimization would require that all financial transactions are made digitally and linked to a centralized national data infrastructure. Currently, even countries with highly digitized economies and high institutional trust (like Sweden) fail to meet that bar, and most countries fall well short of it. Other challenges include how to define the goal function, how to design, train and validate the AI system, and how to build adequate safeguards into the system. Finally, linking and analysing large social scale data across multiple sources is bound to come into tension with privacy regulations, as the controversy surrounding SyRI, the Dutch welfare detection fraud system discussed in Section 1, illustrated.

---

[8] Our thought experiment hinges on data availability, which in turn depends on the maturity of different tax systems and data privacy regulations. However, to some degree, the required data records already exist in the US, Sweden, and elsewhere. Thus, what is needed is not so much a leap in state capabilities as an upgrade of existing infrastructure.

[9] Our line of reasoning is close to Steinmo's (2018) efforts to nudge taxpayers' behaviour via policy experiments and institutional reforms. The difference is that, in our thought experiment, both policy experiments and adjustments are conducted by AI systems.

That said, the idea of using AI systems to optimize tax policy is not just a hypothetical proposition (as in this paper) but a growing field of research at the intersection of economics and computer science. In an article titled *Optimal taxation and insurance using machine learning*, Kasy (2018) proposes a framework that resembles our thought experiment. By combining insights from optimal policy theory and statistical decision theory, Kasy shows how AI systems can perform (quasi-) experiments on taxpayers and draw on this experimental evidence to iteratively choose the policy that maximizes social welfare. The details of Kasy's proposal concern us less here than his conclusion, which is that AI-driven policy optimization "leads to tractable, explicit expressions characterizing the optimal policy choice" and that this "points toward a large area of potential applications for machine learning methods in informing policy."

Kasy's conclusions are echoed by Zheng et al. (2022). In an article titled *The AI Economist,* Zheng et al. argue that "the challenge with policy design comes from the need to solve highly nonstationary decision-making problems where all actors (both taxpayers and the government) are learning." The solution, they suggest, is to design an AI system that uses a two-level RL framework in which both taxpayers and social planners adapt their behaviour and policies to maximize their respective goal functions. Zheng et al. use simulation to show that an AI system based on "two-level RL can find policies that yield higher social welfare than standard baselines" and that AI-driven policy optimization "can be useful without the need for human-coded, application-specific rules." Like our thought experiment, Zheng et al. employ AI systems based on RL to optimize taxation policy. The difference is that while they limit the use of AI systems to simulations that inform policy, our thought experiment envisions an automated implementation of the (continuously) shifting optimal policy.

The findings presented by Kasy (2018) and Zheng et al. (2022) indicates that the technical limitations associated with our thought experiment may be overcome by further research. In contrast, the political limitations may seem insurmountable. Our thought experiment builds on the (unrealistic) assumptions that there is a political consensus to minimize inequality and that technocrats have the freedom to implement the necessary policies. Yet this limitation has little bearing on our overarching argument: we do not propose that AI-driven policy optimization *should* be implemented, nor do we seek to provide a road map for *how* it could be done. Instead, the purpose of our thought experiment is to highlight the normative tensions to which AI rationalization gives rise. We can now turn to these.

## 5. Freedom, equality, and self-determination in the iron cage

Our analysis has shown that building a machine-like tax system via a bureaucratic state is conceivable. However, as this section explains, such a use of AI systems would surface and intensify the social and ethical tensions inherent in Weberian rationalization.

To begin with, using an AI system to optimize tax policy to achieve a predefined normative end – like maximizing economic equality – comes at the expense of other, competing values. Take individual freedom as an example. Both freedom and equality are central ideals of the political ideologies in modern democracies. However, as Charvet (1981) demonstrates, there is a tension between freedom and equality, or between the value of (free) self-determining individuals and the (equal) value that each member of the ethical community accords to the other. Charvet argues in a Kantian vein that individuals must be valued for themselves, for their own ends, to constitute an object of value. But in modern liberal democracies, this individual value must be valued equally for all persons.

Consequently, there is a stand-off between equality and freedom that cannot be reconciled since "equal value" and "free individual self-determination" depend on each other.

The core – Kantian – idea is that the value we place on each other must not be instrumental, whereby we treat each other as means. Yet, that is precisely what an AI-driven tax policy optimization does. To maximize equality, the AI system would not only adjust the tax policy to extract the maximum amount of resources from different taxpaying individuals or groups, it would also learn their behaviours through continuous experimentation – and use that knowledge to shape their future attitudes and preferences. The tension here is that AI rationalization increases centralized control, which is at odds with individual autonomy in relation to the individual's resources. It is important to note that, in our case, this instrumentalization of people applies only to one part of peoples' lives (taxation). Weberian fears of disenchantment may thus be overblown; it is still possible to value people for reasons other than the resources they provide. However, given that taxpaying is a major part of citizens' political life – and likewise a major part of the state's efforts to legitimize legal authority – the rules governing taxation play a large role in people's sense of fairness.

Note that AI rationalization – as envisioned in our thought experiment – is not incompatible with ideas of justice per se. The aims of AI-driven tax policy optimization are consequentialist and seek a redistribution of resources in line with values that have been democratically agreed upon. Rather, the problem is that this control of resources is centralized and hypostatized to the exclusion of other competing political values and overrides citizens' sense of their (non-instrumental) obligations to each other. These tensions sit at the heart of rationalization and have been articulated in different ways by social thinkers from Weber onwards. Yet AI rationalization gives rise to a further normative tension: in theory, the limit of rationalization is that science can only provide the means, it cannot dictate the ends (Cantwell-Smith, 2019). But insofar as AI systems are used to shape the behaviours and preferences of citizens, they de facto shape ends.

It is worth stressing that this problem is not unique to AI-driven tax policy. The more general problem is that new technologies and scientific discoveries go against the free will or autonomy of ethical decision-making. How technology or the disenchantment of the world by science undermines Kantian ethics in a Weberian vein has been noticed, among others, by Gellner in *Legitimation of Belief*. According to Gellner (1975:187), the solution is to grant ourselves "a partial exemption from this cold world since we imposed the order on the world in the first place." This is also what Weber meant when he counterposed "value" rationality to "instrumental" rationality. Value rationality is the idea that human choices should be exempt from being instrumentalized (Brubaker 1984), and it is also why there is such resistance to the idea of technological determinism, whereby it is thought that humans are made subject to impersonal forces. One way to overcome this dilemma is to make the implementation of new technologies subject to democratic acceptance, thus aligning the technological system to human values or needs.[10]

A "meta" reflection can be made here. AI could help facilitate with democratic deliberation in aggregating the inputs of democratic decision-making about what ends technologies should be employed for. In this case, the "value" that is being optimized for is democracy, whether this is conceptualized as "one person, one vote" (each person's value to be valued equally) or in some other way (such as the aggregation of those values in representatives or interest groups). In any event, there is an uneasy "truce" here which leads to many ethical and social quandaries. What is new for our thought experiment is that the use of AI systems to rationalize taxation directly affects the main way in which citizens are bound to each other with respect to the major ethical and political principles embodied in the state. And, if an AI-driven tax policy was applied, it could fundamentally reshape those principles.

---

[10] It should be noted, however, that democratic alignment is no panacea, since the imposition of majority rule on the wills of those who are subject to it can be experienced as a different form of impersonal "determinism".

This is a good place to restate two key points. The first is that AI-driven policy optimization would merely be an intensification of ongoing rationalization processes, which are likely to continue even without this specific application of AI systems. The second is that, as with other technologies, the implementation of AI rationalization is likely to be complex and opaque and, hence, not readily understood by the public. Yet the very idea of AI rationalization, of applying such a "cold", impersonal order to human relations, even if it pertains to the population level rather than to individuals, would likely be regarded as a cause for ethical concerns. While such concerns might be misconceived, the reality, as Weber would have diagnosed it, of a disenchanting technology displacing self-determining beings, and in this case the ethical grounding of the political community of citizens, needs to be addressed in the design and application of AI systems in the public sector.

So far, we have focused on highlighting the normative tensions associated with AI rationalization. However, there are also ways in which AI rationalization can benefit the institutions and processes of liberal democracy. In our thought experiment, the technical details of the workings of the AI-driven tax system may be beyond the public's grasp. However, the ethical principles of egalitarian redistributions are not – and they can be formulated with greater clarity and pursued more rigorously by AI rationalization than legacy policymaking techniques. This would surface the tensions between competing normative visions and enable a discourse about what kind of society citizens want to live in and what trade-offs they are willing to make in the process. For example, one alternative to the egalitarian utilitarian interpretation of equality would be one that maximizes equal opportunities, perhaps in a Rawlsian way, whereby the veil of ignorance (Rawls, 1971) is made transparent instead, or in line with Sen's (1993) capabilities approach. In short, the advantage of AI-driven policy optimization is that it requires that normative ends are made explicit and formalized, thereby subjecting them to public scrutiny and debate.

Our discussion in this section can be summarised as follows. AI-driven policy optimization is indeed within the realm of possibility. In fact, the use of experiments to find the most effective policies and computers to execute those policies efficiently and consistently would merely be a continuation of existing rationalization and bureaucratization processes. However, as our thought experiment has shown, AI rationalization highlights several normative tensions that are often overlooked in the existing literature. Specifically, AI-driven policy optimization can come at the exclusion of other competing political values and may not only override citizens' sense of their (non-instrumental) obligations to each other but also undermine their self-understandings as self-determining beings.

Considered in isolation, each of these tensions has been discussed by different scholars in different contexts. The tension stemming from the incompatibility of different values was examined, among others, by Isaiah Berlin. In his essay *The pursuit of the ideal*, Berlin (1988: 10) writes that "Values can clash. They can be incompatible between cultures, between groups within the same culture, and between you and me. Both liberty and equality are among the primary goals pursued by human beings. But total equality demands the restraint on liberty." Similarly, the tensions concerning the optimization of social systems were noted already by Norbert Wiener. As he wrote in *The Human Use of Human beings* (Wiener, 1954, p212): "The machine, which can learn and can make decisions on the basis of its learning, will in no way be obliged to make such decisions as we should have made, or will be acceptable to us." Finally, the tensions that stem from AI systems shaping human preferences in ways that undermine notions of humans as self-determining beings has been documented by sociologists (Zuboff, 2015) and economists (Thaler & Sunstein, 2008) alike.

What is novel in our exposition is that we have shown that these tensions cannot be simply overcome by building AI systems that are legal, ethical, and safe. There are of course social and ethical challenges that can be addressed through transparency obligations and rigorous engineering practices (Thomas et al., 2023). This is especially true for harms resulting from AI systems failing to operate as advertised. However, it is worth noting that transparency and instrumental rationality build on the very same assumptions that fuel Weberian rationalization processes. Consequently, proposals to design and deploy AI systems in ways that are "fair, transparent, and accountable" risk exacerbating rather than alleviating the tensions associated with AI-driven policy optimization discussed in this section.

## 6. Conclusion

In this paper, we have argued that the social and ethical challenges associated with the use of AI systems in the public sector are best understood against a longer-term trajectory whereby calculability is increasingly imposed on social processes. This claim is based on the observation that computation is a ubiquitous phenomenon, not one unique to AI systems. In fact, Weber defined "rationalization" as the processes whereby traditions, in so far as they serve to guide actions, and random chance are replaced with instrumental rationality, i.e., the most calculable, predictable, and efficient ways of achieving any given objective. And that is precisely what AI-driven policy optimization does.

Foregrounding the continuity between AI-driven policy optimization and larger rationalization processes has two direct implications for the contemporary policy discourse: first, that the social and ethical challenges associated with AI systems intensify ongoing rationalization and bureaucratization processes are distinct from those resulting from AI systems failing to work as advertised; and, second, these two distinct sets of social and ethical challenges call for different remedies, which sometimes stand in tension with each other. Let us consider these in turn.

To start with, our thought experiment with AI-driven tax policy optimization highlights normative tensions that have hitherto received comparatively little attention from policymakers and researchers alike. The literature on fairness, accountability, and transparency in ML focuses on identifying and mitigating harms that result from AI systems failing to perform as intended. For example, AI systems may discriminate against specific individuals and groups (Mehrabi et al, 2021), cause privacy breaches by leaking sensitive data (Narayanan & Shmatikov, 2016), and cause material harm due to malfunctioning or misaligned specifications (Sherer, 2015). These are indeed pressing challenges that demand rigorous treatment from technology providers and policymakers alike. However, as our thought experiment – using AI systems to optimize tax policy to promote social and economic equality – shows, there are other normative tensions that result not from AI **systems** failing to work as intended but that they intensify ongoing rationalization and bureaucratization processes. These tensions include that AI-driven policy optimization (i) comes at the exclusion of other competing political values, (ii) overrides citizens' sense of their (non-instrumental) obligations to each other, and (iii) undermines the notion of humans as self-determining beings. What unites these tensions is that they do not, even in principle, hinge on any malfunctioning or misspecification. Instead, they are likely to be exacerbated as AI systems become increasingly scalable and capable.

Further, emphasizing the link between AI-driven policy optimization and Weberian rationalization suggests that there are limits to what can be achieved through transparency obligations and robust engineering practices. Some of the social and ethical risks AI systems pose can indeed be addressed through proactivity in design. For example, a study by Buolamwini and Gebru (2018) found that gender classification systems were less accurate for darker-skinned females than lighter-skinned males. After being confronted with these findings, technology providers speedily improved the accuracy of their AI systems. This suggests that the problem was not intrinsic but rather resulted from inadequate software development processes and risk management systems, or lack of 'guardrails'. Similarly, other harms that result from AI systems failing to perform as intended can be addressed by ensuring that such systems are used in ways that are legal, ethical, and safe, i.e., by increasing the level of control.

In contrast, the normative tensions stemming from rationalization are not only alleviated but can also be exacerbated when the level of control is increased. This is because the drive towards ensuring that AI systems are designed and used in ways that are fair, transparent, and accountable is based on the same assumptions that fuel rationalization and bureaucratization processes in the first place. These assumptions include the modern ideas that (i) science can sweep away oppressive legal systems and economic policies and replace them with a rule of reason that would rescue humans from moral injustices and (ii) societal processes are calculable and quantifiable (Watson & Mökander, 2023). In fact, building AI systems that are fair, transparent, and accountable is fully in line with the rationalistic ideal of finding the most calculable, predictable, and efficient way of achieving any given objective. However, doing so also propels the disenchantment of the natural world and imposes a cold, impersonal logic to social relationships. This is the essence of Weber's "iron cage."

Before concluding, let us return to the argument put forward by Sparks and Jayaram (2022) in *Rule by Automation*. They right in observing that the use of AI systems in the public sector constrains the space for human discretionary power in the execution of established rules, and thus reduces decision subjects' dependence on arbitrariness and chance. They are also right in concluding that rule by automation should therefore be supported for the same reasons as the rule of law. However, the impact of our formal laws and policies on society and its members has hitherto been cushioned by their inconsistent application and enforcement that results from chance and the discretion exercised by human decision-makers at various levels of the bureaucratic hierarchy.

What our thought experiment with AI-driven policy optimization suggests is that, were our formal laws, policies, and procedures all pursued rationally, i.e., in the most calculable, predictable, and efficient way possible, we could find that we are not so comfortable with them in the first place. This point mainly serves to highlight, however, that making more explicit the aims with which technology can serve us would be a valuable service that AI-driven optimization can provide, if combined with the appropriate guardrails. In the case of tax policies, these guardrails would need to ensure that the aims of policy, democratically arrived at, are implemented in a consistent and transparent way. And the implementation would depend, as discussed earlier, on state capacity, both in the sense of a capable apparatus for gathering information and a bureaucratic organization that works in a transparent and fair – impersonal - way.

To conclude, some of the hardest social and ethical challenges surfaced by AI systems are not unique to computer-centric information processing but mirror tensions at the heart of larger rationalization processes. Tackling these tensions are not only about ensuring that AI systems are designed and deployed in ways that are "legal, ethical, and safe" but also – and perhaps primarily – about confronting hard questions about what (types of) criteria, motivations, and evidence are to be considered legitimate (or at least socially acceptable) for different (private and public) decision-making processes.

**Acknowledgements**

We are grateful to two very helpful anonymous reviewers for this journal, who helped us to clarify and add several key points.

This is a preprint: The manuscript has been accepted for publication in *Social Science Computer review.*_